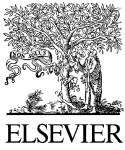

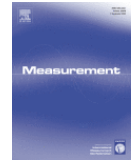

# Real-Time Human Activity Recognition on Edge Microcontrollers: Dynamic Hierarchical Inference with Multi-Spectral Sensor Fusion

Boyu Li[a], Kuangji Zuo[a], Lincong Li[b], Yonghui Wu[b,*]

[a] School of Electrical and Electronic Engineering, Nanyang Technological University, Singapore

[b] School of Physics and Electronics, Henan University, Kaifeng, China

**Abstract**

The demand for accurate on-device pattern recognition in edge applications intensifies, yet existing approaches struggle to reconcile accuracy with computational constraints. To address this critical challenge, a resource-aware hierarchical network based on multi-spectral fusion and interpretable modules, namely the Hierarchical Parallel Pseudo-image Enhancement Fusion Network (HPPI-Net), is proposed to enable real-time, on-device Human Activity Recognition (HAR) tasks. Deployed on ARM Cortex-M4 MCU for low-power real-time inference, HPPI-Net achieves 96.70% accuracy while utilizing only 22.3 KiB of RAM and 439.5 KiB of ROM after optimization. HPPI-Net employs a two-layer architecture: the first layer extracts preliminary features using Fast Fourier Transform (FFT) spectrograms, while the second layer selectively activates either a dedicated module for stationary activity recognition or a parallel LSTM-MobileNet network (PLMN) for dynamic states. PLMN fuses FFT, Wavelet, and Gabor spectrograms through three parallel LSTM encoders and refines the concatenated features with Efficient Channel Attention (ECA) and Depthwise Separable Convolution (DSC), thereby offering channel-level interpretability while substantially reducing multiply–accumulate operations. Compared to MobileNetV3, HPPI-Net notably increases accuracy by 1.22% and significantly reduces RAM usage by 71.2% and ROM usage by 42.1%. These results demonstrate that HPPI-Net realizes a favorable accuracy-efficiency trade-off and provides explainable predictions, establishing a practical solution for wearable, industrial, and smart-home HAR on memory-constrained edge platforms.

\* Corresponding author at: School of Physics and Electronics, Henan University, Kaifeng 475001, China

*E-mail address:* defey@henu.edu.cn.

## 1. Introduction

The proliferation of wearable sensing platforms and the growing demand for on-device intelligence have led to a pressing need for real-time, accurate human activity recognition (HAR) on edge microcontrollers [1,2]. Applications ranging from healthcare monitoring [3] and sports analytics to smart factory environments increasingly require embedded HAR systems capable of running under tight resource budgets while ensuring low latency and high reliability [4,5]. However, reconciling recognition performance with the extreme limitations of memory, computation, and energy typical of microcontroller-class hardware remains a core research challenge [6,7].

Conventional deep learning-based HAR models typically rely on single time-frequency transformations [8] such as the Fast Fourier Transform (FFT) [9], Wavelet Transform (WT) [10], or Gabor Transform (GT) to generate input spectrograms [11]. While each of these methods captures specific signal characteristics (periodicity, transient features, or localized energy), they struggle independently to represent the full complexity of both dynamic and static human motion patterns [12,13]. Moreover, models that attempt to fuse multiple spectral features often do so through monolithic architectures with high resource overheads, rendering them unsuitable for real-time execution on embedded systems [14,15].

To bridge this gap, we propose the **Hierarchical Parallel Pseudo-Image Enhanced Fusion Network (HPPI-Net)**, a novel model explicitly designed for resource-constrained inference without compromising recognition accuracy [16-18]. HPPI-Net leverages a modular, two-stage hierarchical inference strategy. An ultra-lightweight first layer quickly coarsely classifies activity states (Moving, Stationary, Cycling) based on FFT-derived spectrograms [19]; subsequently, a second stage activates specialized branches for fine-grained recognition according to initial classification results. Dynamic activities invoke a **Parallel LSTMs-MobileNet Network (PLMN)**, which fuses FFT, WT, and GT pseudo-images through lightweight attention and Depthwise Separable Convolution (DSC) [20]. Meanwhile, stationary activities are classified by a low-overhead FFT-based CNN-LSTM module reused from the first stage [21].

PLMN is the core innovation in HPPI-Net. It integrates FFT, WT, and GT spectrograms via three parallel LSTM encoders, followed by Efficient Channel Attention (ECA) and DSC modules. The ECA block enables efficient channel-level feature recalibration while preserving interpretability, providing insight into the contribution of each spectral branch and sensor axis [22,23]. The use of DSC, originally introduced in MobileNet architectures, significantly reduces parameter count and Multiply-Accumulate Count (MACC) [24,25]. Importantly, HPPI-Net does not adopt the full MobileNet structure, but rather selects its most deployment-friendly components for maximal balance between accuracy and resource utilization [26].

In addition, HPPI-Net supports explainable inference through post-hoc MLP-based analysis that quantifies the contribution of different feature branches and sensor channels [27]. This design makes the model more transparent and trustworthy for use in healthcare or safety-critical applications.

The main contributions of this paper are summarized as follows:

1) **Resource-Aware Hierarchical Inference Framework:** HPPI-Net enables conditional invocation of computationally intensive modules, significantly reducing dynamic RAM usage. The hierarchy also ensures architectural modularity, which is highly suitable for embedded system integration.

2) **Spectral-Channel Fusion with Lightweight and Interpretable Modules:** The dynamic recognition path integrates FFT, WT, and GT features using three parallel LSTM encoders. These features are adaptively fused through ECA and processed by DSC blocks. This combination offers both computational efficiency and interpretable attribution to specific spectral domains.

3) **Microcontroller Deployment with Full-Stack Optimization and Explainability:** HPPI-Net is deployed on ARM Cortex-M4 MCU, achieving the optimal balance between accuracy and limited resources, and realizing real-time HAR inference. Meanwhile, a lightweight MLP-based post-hoc analysis provides feature- and axis-level interpretability, reinforcing its trustworthiness in real-world edge applications.

An overview of the overall system architecture is depicted in **Fig. 1**.

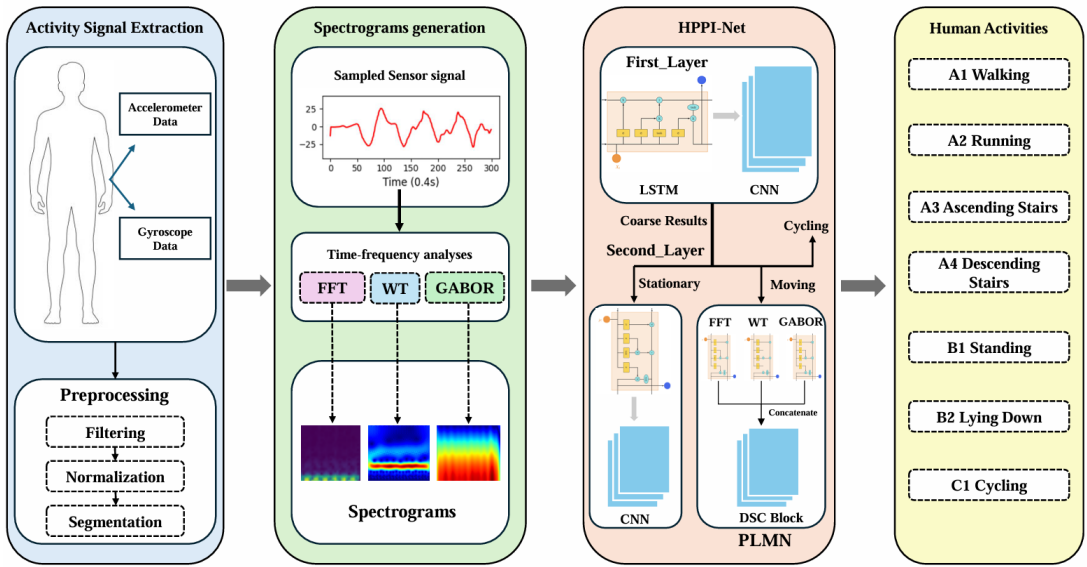

**Fig. 1.** Time-domain characteristics across 6-axis data.

## 2. Background

### 2.1. Sensor-Based Human Activity Recognition

Sensor-based HAR leverages data from inertial measurement units (IMU), typically including accelerometers and gyroscopes, to classify human activities in various fields such as healthcare, sports analytics, and interactive applications [28,29]. Conventional HAR systems commonly employ machine learning methods such as Support Vector Machines (SVM) [30] and Random Forests (RF) [31]. Recently, deep learning methods, notably Convolutional Neural Networks (CNN) [32] and Long Short-Term Memory (LSTM) networks [33], have significantly advanced HAR performance due to their powerful representation learning capabilities. Additionally, lightweight architectures such as MobileNet [34] have gained attention due to their suitability for resource-constrained edge devices.

However, recent approaches based on CNN and LSTM have achieved remarkable accuracy, their end-to-end architectures often fail to capture structural sparsity, modularity, or interpretability—critical factors for deployment on constrained microcontroller units (MCU) [35]. Furthermore, these models typically treat all sensor axes and features uniformly, overlooking axis-specific dynamics and redundancy [36].

The integration of wearable HAR device becomes a reality by setting up the model on the edge device. Descriptions of the activities are summarized in **Table 1**.

**Table 1**. Descriptions of the human activities.

| Activity reference | Desciption of activity |
|---|---|
| A | Moving |
| A1 | Walking |
| A2 | Running |
| A3 | Ascending stairs |
| A4 | Descending stairs |
| B | Stationary |
| B1 | Standing |
| B2 | Lying down |
| C1 | Cycling |

## 2.2. Multi-Spectral Feature Extraction

Multi-spectral feature extraction methods, including Fast Fourier Transform (FFT), Wavelet Transform (WT), and Gabor Transform (GT), have been widely adopted for analyzing time-series sensor data due to their robust ability to extract discriminative frequency-domain features.

1) FFT

FFT is extensively used to convert time-domain signals into frequency-domain representations, revealing periodic characteristics within activities [37]. FFT is mathematically expressed as:

$$X(k) = \sum_{n=0}^{N-1} x[n] \, e^{-j2\pi kn/N} \tag{1}$$

where $x[n]$ is the input time-domain signal, $X(k)$ represents the output frequency-domain component at index $k$, $N$ is the total number of samples.

2) WT

WT provides localized analysis in both time and frequency domains, capturing transient features effectively. Discrete Wavelet Transform (DWT) using Haar wavelets [38] can be expressed as:

$$a_j[k] = \sum_n x[n]\Phi_{j,k}[n], \quad d_j[k] = \sum_n x[n]\Psi_{j,k}[n] \tag{2}$$

where $a_j[k]$ and $d_j[k]$ represent the approximation (low-frequency) coefficients and detail (high-frequency) coefficients at scale $j$, index $k$, respectively. $\Phi_{j,k}[n]$ and $\Psi_{j,k}[n]$ denote the scaling function (Haar basis) and wavelet function (Haar wavelet basis), respectively. $j$ is the scale level and $k$ is the translation index.

3) GT

GT leverages Gaussian window functions to achieve optimal time-frequency localization. Its continuous form is defined as [39]:

$$G(t,f) = \int_{-\infty}^{+\infty} x(\tau) \cdot e^{-\pi(\tau-t)^2/\sigma^2} \cdot e^{-j2\pi f\tau} \, d\tau \tag{3}$$

where $G(t, f)$ represents the Gabor Transform coefficients, $x(\tau)$ is the discrete-time sensor signal, and $\sigma$ is the standard deviation of the Gaussian window, controlling the trade-off between time and frequency resolution.

In this study, each transform method is applied to six sensor channels: three-axis accelerometer and three-axis gyroscope data. Each channel generates an independent spectrogram, capturing the time-frequency characteristics of the data. The spectrograms of the six channels are then stacked along new dimensions to form a three-dimensional array (time, frequency, channel) that serves as an input to the classification model.

In practice, different transformations exhibit axis-specific sensitivity: for instance, GT often emphasizes periodic patterns prevalent in gyroscope-z signals during cycling, while WT responds to transient bursts seen in acceleration-y during stair climbing. The fusion of FFT, WT, and GT allows complementary exploitation of low-, mid-, and high-frequency components, providing a richer time-frequency representation essential for accurate recognition of both dynamic and stationary activities.

### 2.3. Lightweight Deep Learning Architectures

Deploying deep learning models on embedded systems requires consideration of computational efficiency. Lightweight architectures, such as MobileNet, employ Depthwise Separable Convolution (DSC), significantly reducing computational costs compared to standard convolutional layers [24]. In this approach, a spatial convolution is applied independently to each input channel (depthwise), followed by a pointwise 1×1 convolution to combine the outputs. This structure is defined as:

$$Conv_{dsc}(X) = Conv_{k \times k}^{depthwise}(X) + Conv_{1 \times 1}^{pointwise}(X), \ X \in \mathbb{R}^{H \times W \times C_{in}} \ , \ Conv_{dsc}(X) \in \mathbb{R}^{H \times W \times C_{out}} \tag{4}$$

where $X$ and $Conv_{dsc}(X)$ represent the input feature map and output feature map, respectively. $Conv_{k \times k}^{depthwise}$ denotes a depthwise convolution of kernel size k×k applied independently to each input channel, and $Conv_{1 \times 1}^{pointwise}$ is a pointwise (1×1) convolution that mixes channel information. Compared with traditional convolution (with cost of $k^2 \cdot C_{in} \cdot C_{out}$), the DSC reduces the computation to $k^2 \cdot C_{in} + C_{in} \cdot C_{out}$, making it suitable for microcontroller deployment.

In this work, instead of adopting the full MobileNet structure, only the core DSC modules are utilized to construct lightweight convolutional blocks for the post-attention fusion stage, enabling deployment feasibility and real-time inference on resource-constrained embedded devices.

In addition, Efficient Channel Attention (ECA) [22] is adopted to enhance channel-level feature interaction without significantly increasing computational burden. ECA computes attention weights by performing a one-dimensional convolution on the globally pooled feature vector:

$$ECA\,(x) = \ \sigma(Conv1D\,(GAP(x))), x \in \mathbb{R}^{C \times H \times W}, GAP(x) \in \mathbb{R}^{C} \tag{5}$$

where $x$ is the input feature map, $GAP\,(x)$ represents the global average pooled vector, σ denotes the sigmoid activation function. The resulting attention vector is subsequently used to rescale the original channel features.

Dimensionality reduction is avoided, and local cross-channel interactions are preserved, thereby enabling an efficient yet powerful attention mechanism suitable for resource-constrained edge devices.

Inspired by the MobileNet architecture, the full design is not adopted. Instead, lightweight DSC modules are integrated with an attention-based feature fusion mechanism, tailored for compatibility with microcontroller-based deployment.

## 2.4. Explainability and Edge Deployment

The lack of transparency in deep neural networks limits their applicability, especially in safety-critical or healthcare scenarios where model explainability is essential. To meet the requirements of computational resources, the lightweight attention mechanism is incorporated, and a hierarchical inference structure is adopted, in which complex sub-networks are activated only when necessary. Furthermore, instead of relying exclusively on attention modules such as ECA, whose weights are uniformly distributed, the post-hoc MLP-based feature attribution approach is employed to reveal class-specific contributions from FFT, WT, and GT inputs [36].

For embedded deployment, the generic neural network compilation toolchain is utilized that supports quantization, operator fusion, and memory-aware optimizations. This process converts high-level deep learning models into low-level C code tailored for microcontroller execution, ensuring real-time performance under severe resource limitations.

## 3. Methods

### 3.1. Data Acquisition and Pre-processing

In this study, a single 6-axis IMU is strapped to the dominant wrist of participant and sampled at 50 Hz via the $I^2C$ bus of an STM32F407ZGT6-based wearable device. Twenty healthy volunteers (mean age 22 years, mean weight 65 kg) performed seven daily activities (Walking, Running, Ascending Stairs, Descending Stairs, Standing, Lying Down, and Cycling) in a controlled indoor environment. Raw accelerometer and gyroscope readings are first converted to physical units and denoised by a 3-point median filter to suppress impulsive noise from mechanical vibration and electromagnetic interference. The denoised stream is segmented with a fixed, non-overlapping window of 16 samples, which offers a trade-off between temporal resolution and on-device memory footprint. For each window, the majority label is assigned as its activity class. As illustrated in **Fig. 2**, the six subplots for each activity correspond to the 3-axial accelerometer (ax, ay, az) and the tri-axial gyroscope (gx, gy, gz) data.

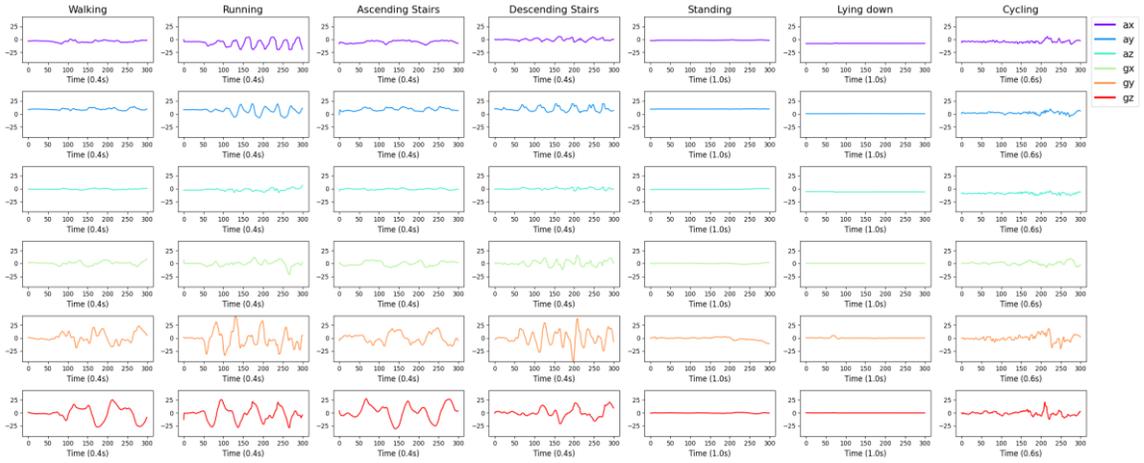

**Fig. 2.** Time-domain characteristics across 6-axis data.

Clearly, all types of activity exhibit periodicity. Walking and Running show pronounced periodic oscillations in both accelerometers (ax, ay) and gyroscopes (gz), reflecting the regular high-frequency movements of the wrist during these strenuous activities. On the other hand, fluctuations in all axes for static activities such as Standing and Lying down are minimal and remain almost constant in value. Finally, periodic gyroscopic activity occurs when Cycling, especially along the z-axis (gz), suggesting rhythmic wrist movements during Cycling.

### 3.2. Multi-Spectral Pseudo-Image Generation

In this study, to comprehensively capture the spatio-temporal characteristics of sensor signals, three time-frequency representations (FFT, WT and GT) are generated and compared under specific configurations. Each of these methods provides a unique perspective on the frequency and temporal properties of the signals. Each transformation method applies a sliding window of length 16 along the time axis, and the output dimension is (16, 6), with 16 representing the length of the time dimension and 6 representing the six dimensions of the signal (3-axial acceleration and 3-axial angular velocity).

As shown in **Fig. 3**, the subplots clearly demonstrate the distribution of these time-frequency representations under various activities, revealing the response characteristics of each transformation method to different human activities.

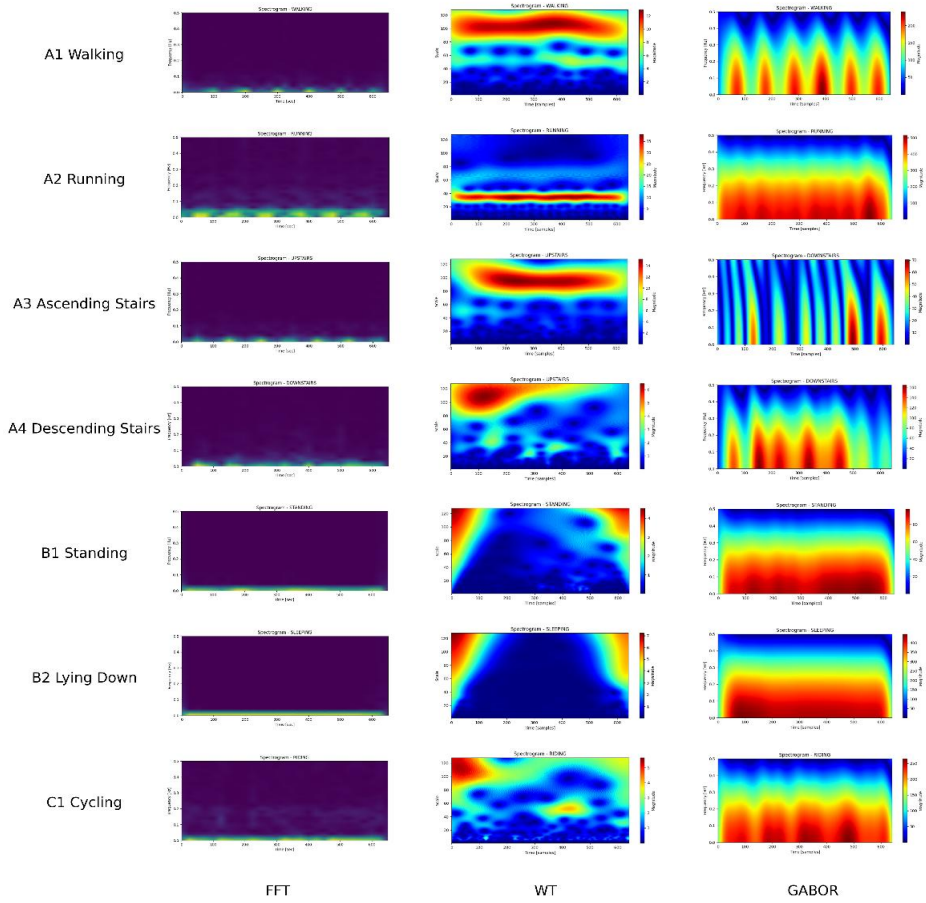

**Fig. 3.** Spectrograms of three time–frequency analyses for seven activities.

1) FFT

This time-frequency representation method, based on fixed time windows, effectively captures relevant features for broad classifications of activities Moving (A), Stationary (B), and Cycling (C). However, due to the low temporal resolution, FFT presents blurred frequency boundaries. Its performance is less satisfactory for non-periodic activities and situations with rapid frequency changes, such as A3 and A4.

2) WT

The multi-resolution analysis capability of WT demonstrates high resolution in both time and frequency domains, especially excelling at capturing transient changes during complex dynamic activities. In the spectrograms of A3 and A4, the higher local feature resolution of WT is evident, with these features predominantly concentrated in the low-frequency region, indicating that these activities primarily involve overall trunk movement. Furthermore, in activities B1 and B2, WT distinctly differentiates between high-frequency and low-frequency signals, showcasing its advantage in localizing static activities over time.

3)   GT

The Gabor Transform combines the advantages of high time and frequency resolution to effectively capture the localized features of signals in time and frequency simultaneously. GT provides frequency distribution information in a short time window and is particularly suitable for capturing activities with strong periodic and transient variations such as (A1 and A2).

Combined with the above analysis, the FFT has an advantage in capturing features quickly, the WT is good at transient feature detection, and the GT is more effective in capturing periodic features. By integrating these features, the model can effectively utilize the rich spatio-temporal features in human activity data.

*3.3. Hierarchical Inference Framework*

To achieve the robust trade-off between classification performance and computational efficiency on resource-constrained microcontrollers, the state-of-the-art Hierarchical Parallel Pseudo-Image Enhanced Fusion Network (HPPI-Net) is proposed. HPPI-Net consists of two cascaded stages: an initial coarse-grained classifier for activity state discrimination, and a fine-grained classifier tailored to the inferred motion category. As shown in **Fig. 4**, this design enables selective model invocation, where heavier architectures are only activated when necessary, thereby minimizing redundant memory and compute overhead.

1)   First_Layer Classification: Coarse-Grained Activity Recognition

As shown in **Fig. 4(a)**, a lightweight CNN–LSTM model is deployed in the first layer to perform coarse-grained classification into A, B, and C. FFT spectrograms are adopted as input, as they efficiently reveal activity-specific frequency patterns while maintaining fixed dimensions suitable for lightweight convolution. Specifically, a two-layer 2D convolutional network with batch normalization and max pooling is used to extract salient spatial features from pseudo-images. These features are then condensed via global average pooling and passed into a single-layer LSTM module to capture temporal dependencies across frames. Finally, a dense softmax layer generates a probability distribution over the three coarse-grained classes. Upon completing the initial classification, the second layer classifier is responsible for fine-grained classification. Based on the initial classification results, the system selects different models to process specific activities.

2)   Fine-Grained Classification for Motion Activities

When the coarse classifier designates a segment as Moving (A), the framework transitions to a specialised branch that undertakes fine-grained recognition without interrupting real-time execution. This branch, termed the Parallel LSTM-MobileNet Network (PLMN), is conceived to reconcile two normally conflicting goals: the expressive power needed to separate subtle motion classes and the stringent Random Access Memory (RAM)/ Read-Only Memory (ROM) ceilings of an MCU.

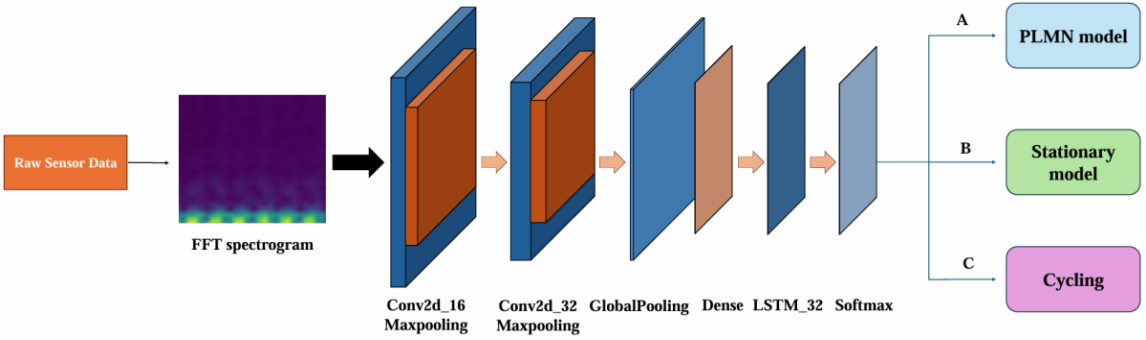

(a)

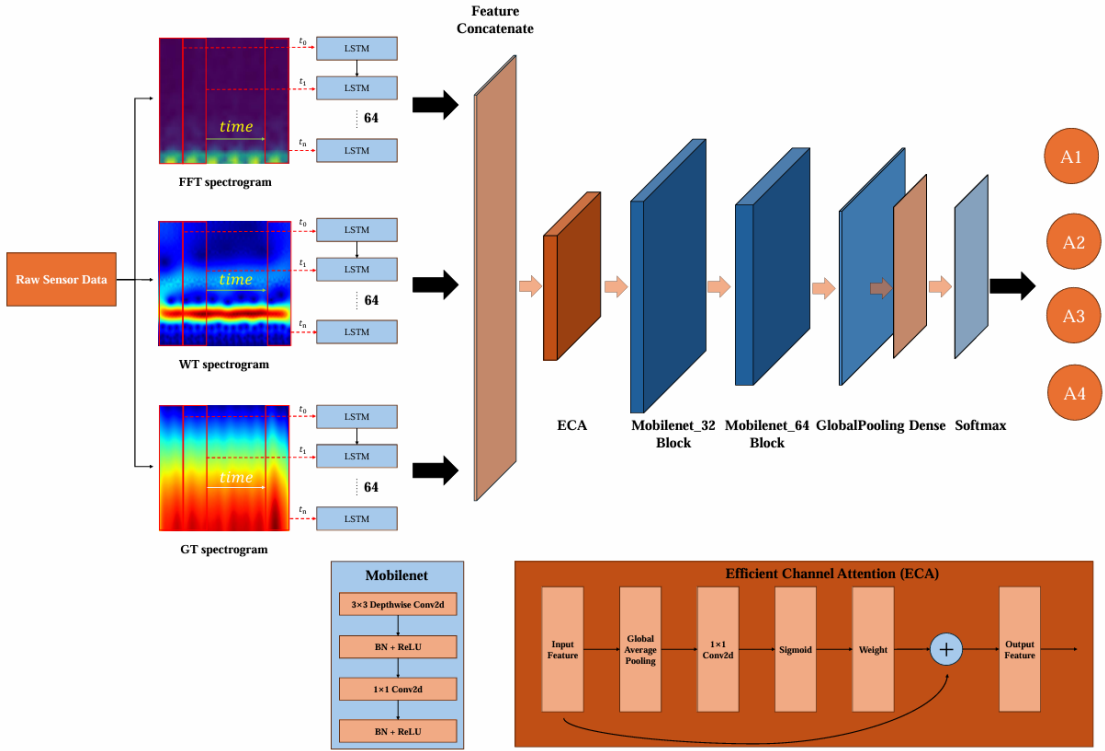

(b)

**Fig. 4.** Frame diagram of HPPI-Net Structure. (a) First_Layer network; (b) PLMN.

The structure of PLMN is shown in **Fig. 4(b)** and leverages three parallel LSTM to process pseudo-images derived from different signal transformations. The FFT channel focuses on capturing the time-frequency features of the motion signal and analyzing periodic components, the WT channel provides multi-scale analysis of transient signal features, and the GT channel excels at providing time-frequency localized representation and particularly well suited for analyzing rapidly changing signal components. The resulting sequences (shape 16×6) are forwarded to three parallel LSTM encoders, each composed of 64 hidden units. The terminal hidden states are concatenated, yielding a 192-dimensional joint representation.

Then the concatenated representation is refined by an attention-guided depthwise pipeline. A lightweight ECA module applies a 1-D convolution followed by a sigmoid gate, generating per-channel weights that rescale the joint vector. This strategy avoids dimensionality reduction, retains local cross-channel interactions, and adds only a handful of parameters—an attractive property for embedded compilation. The reweighted vector is reshaped to 1×1×192 and passed through two DSC modules inspired by MobileNet. By factorising spatial and channel mixing, each DSC module delivers nearly an order-of-magnitude Multiply-Accumulate Count (MACC) reduction relative to conventional convolutions while preserving accuracy.

A global-average-pooling layer and a four-unit soft-max classifier complete the branch, producing probabilities for A1, A2, A3 and A4. The channel weights emitted by ECA are retained and later interpreted with a lightweight MLP, providing class-level insight into the relative importance of FFT, WT and GT features.

3) Fine-Grained Classification for Stationary Activities

For the classification result to be Stationary (B), the system bypasses PLMN and invokes a simplified classifier tailored to distinguish between Standing and Lying Down. For static activities, time-series variation is minimal, and thus simpler models suffice.

The same FFT-based spectrogram construction of First_Layer classifier is reused. The classification model retains the CNN-LSTM structure but restricts the output to two classes B1 and B2. This sharing of architecture and weights minimizes additional memory load, achieving both architectural modularity and deployment efficiency.

HPPI-Net maximizes the complementary information between different pseudo-images, which significantly improves the accuracy and robustness of the HAR task, even with constrained computational resources.

*3.4. Embedded Deployment Optimisation*

To guarantee real-time inference on resource-constrained microcontrollers, the proposed architecture is accompanied by a set of deployment-oriented design strategies.

1) Resource-Aware Module

HPPI-Net is structured into three modular components: the First-Layer classifier, the Stationary classifier, and the PLMN classifier. Each component is compiled as an independent binary and managed separately during execution. To minimise memory residency, only the First-Layer classifier remains permanently loaded, while the downstream classifier is conditionally loaded at runtime based on the coarse classification output. This runtime partitioning reduces inactive memory overhead and supports power-aware deployment on resource-constrained microcontrollers.

The total ROM and MACC are calculated by summing the values for all components, since each must be stored and compiled prior to deployment. In contrast, the RAM and classification Accuracy depend on the execution path during inference. Let $ACC_i$ and $RAM_i$ denote the Accuracy and RAM of module $i \in \{$FL, PLMN, S$\}$ (First_Layer, PLMN, Stationary). Given the activation probability $p$ that a sample proceeds to PLMN, the expected values of Accuracy and RAM of HPPI-Net are calculated as follows:

$$ACC = ACC_{FL} * \{p * ACC_{PLCN} + (1 - p) * ACC_S\} \tag{6}$$

$$RAM = RAM_{FL} + p * RAM_{PLCN} + (1 - p) * RAM_S \tag{7}$$

2) Adaptive Sampling-Rate Adjustment (ASRA)

To further conserve energy during inference, an adaptive sampling mechanism is implemented based on the coarse-grained activity output. Upon classification by the first-layer module, the IMU sampling rate is reconfigured in real-time. Specifically, stationary states (B1, B2) trigger a low-rate configuration (10 Hz), moderate-intensity dynamics (C) invoke an intermediate rate (25 Hz), and high-intensity movements (A1, A2, A3, and A4) retain the full sampling rate (50 Hz).

*3.5. Explainability and Feature Analysis*

To evaluate the validity of spectral fusion and gain insights into the internal reasoning of model, several post-hoc analyses are proposed to examine how different branches and sensor dimensions contribute to the final decision.

1) Feature Branch Attribution

While the ECA module is used during inference to recalibrate fused features, its per-channel weights often lack sufficient variation to serve as effective explanatory signals. To obtain clearer attribution, a lightweight MLP is trained post hoc to regress from the fused feature vector to the relative importance of the three spectral branches: FFT, WT, and GT. For each test sample, the MLP outputs a three-dimensional weight vector, which is averaged per class to reveal class-specific preferences.

To investigate which sensor axes are emphasized within each spectral domain, ECA weights are further decomposed according to axis assignment (ax, ay, az, gx, gy, gz). By aggregating these weights across classes, class-level axis attention profiles are constructed.

2) Inter-Feature Correlation Analysis

Finally, to validate that multi-spectral fusion captures non-redundant information, pairwise Pearson correlations are computed between flattened FFT, WT, and GT feature vectors. A low correlation implies orthogonality in representation, supporting the use of parallel encoders. The resulting heatmaps confirm that while some channels share global motion trends, the transforms emphasize distinct spectral characteristics overall, thereby justifying the added architectural complexity.

## 4. Experimental Results and Analyses

### 4.1. Experimental Setup

All experiments were conducted on a workstation running Ubuntu 22.04 LTS (64-bit), equipped with an Intel Core i7-14700KF CPU, 32 GB RAM, and an NVIDIA GeForce RTX 2080 GPU. Model implementation and training were performed using TensorFlow 2.18.0 with the Keras API, running on Python 3.12.6.

To ensure fairness and reproducibility, all models were trained under identical conditions. The Adam optimizer was employed with an initial learning rate of 0.0001. Each model was trained for a maximum of 100 epochs with a batch size of 32. For each configuration, an early stopping criterion was applied based on validation loss with a patience of 5 epochs. All preprocessing steps (segmentation, median filtering, and standardisation) were identically applied across models.

For embedded deployment evaluation, models were quantised to 8-bit fixed-point and compiled to ANSI-C using a generic MCU-oriented neural network toolchain. Deployment was performed on an STM32F407ZGT6 microcontroller (ARM Cortex-M4, 168 MHz, 192 KiB SRAM, 1024 KiB Flash).

*4.2. Experimental Results*

By examining the classification accuracy curves and confusion matrices of each component module, the effectiveness of HPPI-Net can be analyzed in handling complex pattern recognition tasks from multiple perspectives.

**Table 2**. HPPI-Net experimental results.

| Network | Accuracy | ROM (KiB) | RAM (KiB) | MACC |
|---|---|---|---|---|
| First_Layer | 99.35% | 210.9 | 87.2 | 142048 |
| PLMN | 95.17% | 890.6 | 25.9 | 889377 |
| Stationary | 99.50% | 210.9 | 87.2 | 142000 |
| HPPI-Net (Before optimization) | 96.70% | 1312.4 | 143.9 | 1173425 |
| HPPI-Net (Afteroptimization) | 96.70% | 439.5 | 22.3 | 1173425 |

**Table 2** summarises the module-level performance of HPPI-Net. The First-Layer, PLMN and Stationary classifiers achieve validation accuracies of 99.35 %, 95.17 % and 99.50 %, respectively. The entry HPPI-Net (Before optimisation) represents the theoretical system footprint obtained by assuming that all three modules are statically resident in memory. Its Accuracy and RAM figures are the expected values derived from the probabilistic activation model of **Section 3.4**, with the branch-selection probability fixed at $p$=0.5. By contrast, HPPI-Net (After optimisation) reports the actual embedded deployment scenario.

During deployment to the ARM Cortex-M4 microcontroller, quantization and pruning are applied to reduce memory usage while preserving classification performance. The deployed HPPI-Net occupies only 22.3 KiB of RAM and 439.5 KiB of ROM, conforming to the on-chip memory constraints of the target device. The total computational load amounts to 1173425 MACC, which is well within the processing capacity of the Cortex-M4 at 168 MHz, thereby ensuring real-time inference capability. The accuracy and confusion matrix of each model is illustrated in **Fig. 5**.

The First_Layer model performs well in distinguishing broad features, with some misclassifications observed between Moving (A) and Stationary (B). For the PLMN model, recognition performance is excellent in high-frequency movements, with minor misclassification only between A3 and A4. The Stationary model achieves exceptionally high accuracy due to the clear distinctiveness of its features.

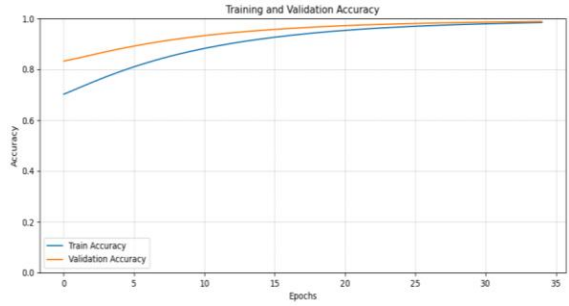

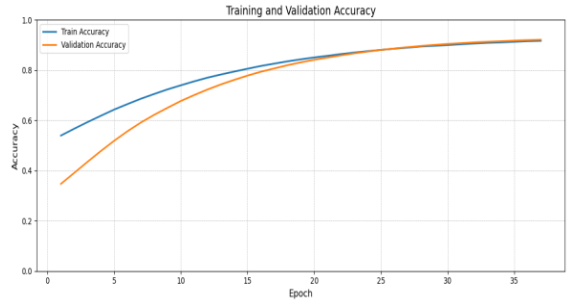

(a)

(b)

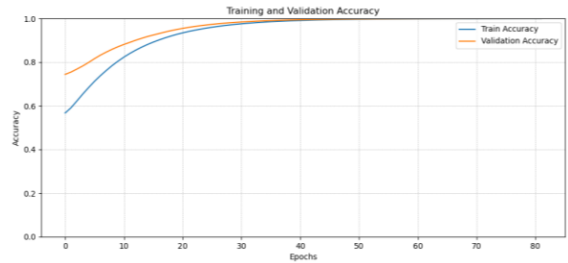

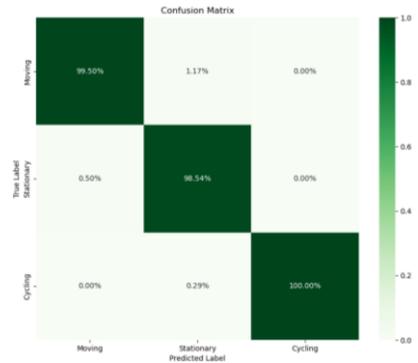

(c)

(d)

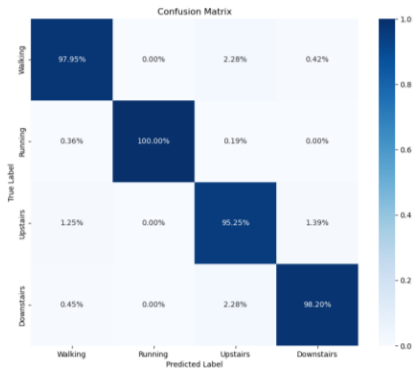

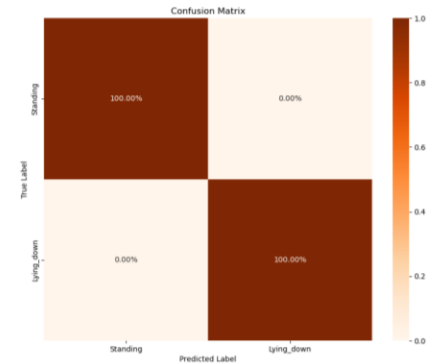

(e)

(f)

**Fig. 5.** Classification accuracy and confusion matrix. (a) First_Layer network accuracy; (b) PLMN network accuracy; (c) Stationary network accuracy; (d) First_Layer network confusion matrix; (e) PLMN network confusion matrix; (f) Stationary network confusion matrix.

*4.3. Explainability Results*

1)  Feature Branch Attribution

As shown in **Fig. 6 (a)** and **Fig. 6 (b)**, the per-class attention contributions of FFT, WT, and GT branches are visualized both as absolute attention weights and normalized percentages. Across all four motion classes (Walking, Running, Ascending, and Descending), GT maintains a consistently higher contribution (Average 34.25%), followed closely by WT. FFT, while slightly lower, remains non-negligible, confirming the complementary role of each transform. These results justify the inclusion of all three spectral branches and demonstrate that the attention mechanism does not collapse to uniform weighting.

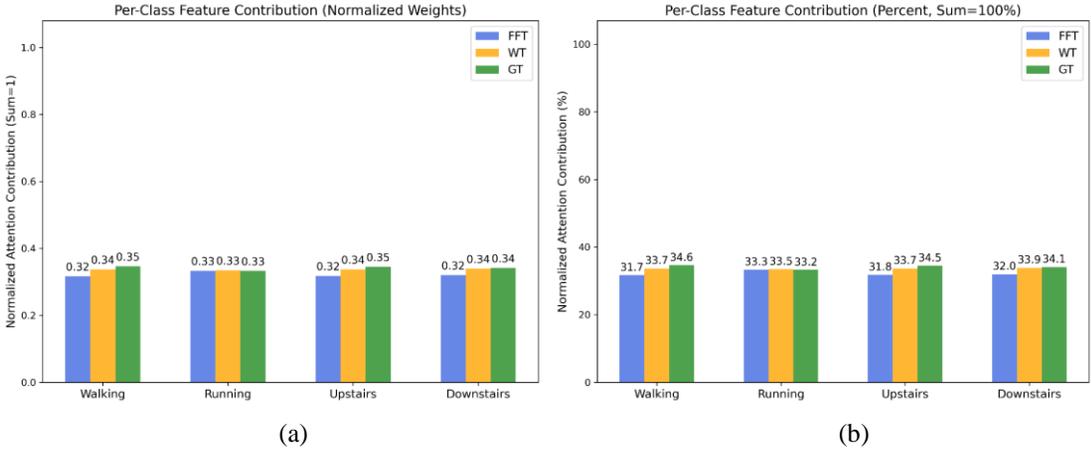

(a)                                                                      (b)

**Fig. 6.** Per-class attention contribution of each branche. (a) Absolute attention weights; (b) Normalized percentages.

2)  Axis-Specific Attention Distribution

**Fig. 7** presents the average attention weight assigned to each sensor axis under different spectral views. Notably, GT assigns the highest weights to gx, gz and gy, indicating strong reliance on gyroscopic input during periodic movements (Moving and Cycling). In contrast, WT emphasizes az, consistent with its sensitivity to short bursts and vertical transitions (Ascending stairs and Descending stairs). This axis-level pattern aligns with biomechanical expectations and supports the physiological grounding of the learned representations.

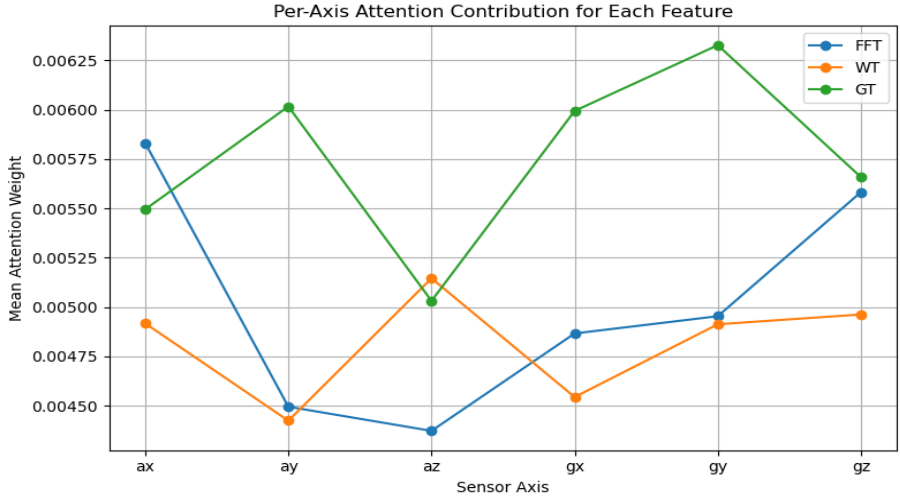

**Fig. 7.** Per-axis attention contribution for each feature.

3)   Feature correlation analysis

To evaluate potential redundancy between branches, the Pearson correlation coefficients among FFT, WT, and GT features were computed and are visualized in **Fig. 8**. The results indicate weak to moderate negative correlation, suggesting that the three domains encode largely non-overlapping information. This provides quantitative support for the fusion design and demonstrates that the added complexity of maintaining three branches is justified by information diversity.

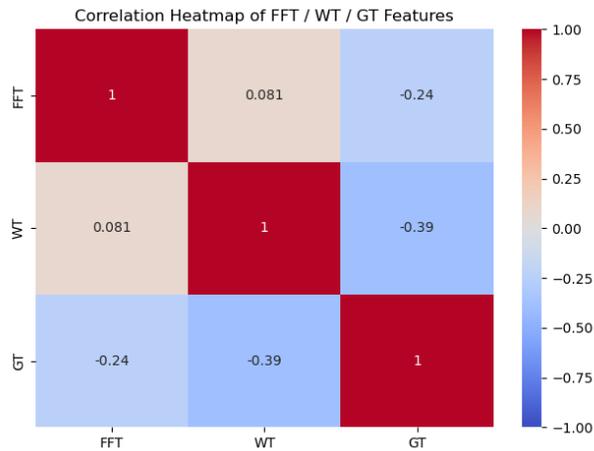

**Fig. 8.** Correlation heatmap among FFT, WT, and GT features.

In summary, the explainability analysis reveals that HPPI-Net attends to semantically meaningful features at both branch and axis levels, and that its multi-branch structure captures distinct, non-redundant spectral perspectives. All analyses were performed offline and do not affect model deployment.

### 4.4. Ablation Study

A set of ablation and comparative experiments are designed to evaluate the impact of spectral diversity, attention integration, and convolutional structure on overall performance. All models follow the same backbone as PLMN without optimization, unless otherwise specified.

**Table 3**. Ablation study of PLMN.

| Network | Accuracy | ROM (KiB) |
|---|---|---|
| FFT | 84.03% | 327.5 |
| WT | 88.24% | 327.5 |
| GB | 89.98% | 327.5 |
| PLMN (no attention) | 93.81% | 900.6 |
| PLCN | 94.68% | 1616.8 |
| PLMN | 95.17% | 890.6 |

**Table 3** highlights the superiority of the proposed PLMN architecture by comparing it against several ablated or structurally modified variants. The table demonstrates that PLMN not only achieves the highest accuracy but also maintains a compact ROM footprint, making it well-suited for embedded deployment.

To begin with, the effectiveness of multi-spectral fusion is examined. The first three rows report the results of PLMN variants that use only a single spectral input (FFT, WT, or GB), while retaining the three-branch structure. Although the PLMN occupies nearly 2.5 times more ROM than the single-input model, this is acceptable due to its improved accuracy of about 10%.

In terms of attention mechanisms, removing the ECA module (PLMN without attention) results in a non-negligible drop in accuracy to 93.81%, despite keeping the rest of the structure unchanged. Notably, the ROM size increases to 900.64 KiB, which may seem counterintuitive but reflects the loss of regularity and structure-aware optimizations that attention mechanisms facilitate during model compilation.

Finally, the PLCN (Parallel LSTM-CNN Network) variant replaces the depthwise separable convolutions in PLMN with standard 3×3 convolutions. As accuracy drops slightly (94.68%), ROM usage nearly doubles to 1616.75 KiB, confirming the efficiency of the MobileNet-inspired design in preserving performance with minimal memory.

In conclusion, PLMN achieves the best balance between accuracy (95.17%) and memory efficiency (890.6 KiB), outperforming all single-feature and structurally simplified alternatives. These results validate the architectural choices of multi-branch spectral fusion, lightweight attention, and depthwise separable convolution as critical to the success of HPPI-Net under embedded constraints.

## 4.5. Comparison with State-of-the-Art Models

To evaluate the effectiveness of HPPI-Net in terms of both recognition performance and resource efficiency, several widely adopted baseline models for sensor-based activity recognition are compared. All comparisons are conducted under a unified experimental pipeline, using the same input segmentation, feature extraction (FFT, WT, GT), and training configuration. The following architectures are selected for comparison, all the State-of-the-Art （SOTA）models here adopt the multi-feature strategy.

- MobileNetV3 [26]: A compact convolutional network originally designed for mobile vision tasks.
- Attention-LSTM [40]: A recurrent model enhanced with temporal attention mechanisms, adapted for multi-spectral inputs.
- Transformer+RF [41]: A hybrid architecture combining a lightweight Transformer encoder with a Random Forest classifier.
- HARMamba [42]: A hardware-aware variant of the SSM-Mamba model, restructured for compatibility with Cortex-M series processors.

**Table 4**. Comparative analysis of HPPI-Net.

| Network | Accuracy | RAM (KiB) | ROM (KiB) | MACC |
|---|---|---|---|---|
| Attention-LSTM | 94.19%[5] | 16894.5[4] | 24983[3] | 14241280[4] |
| HARMamba | 95.05%[4] | **240.2[2]** | **1,204.1[1]** | **856512[2]** |
| MobileNetV3 | 95.48%[3] | 500.0[3] | 2,267.6[4] | **729024[1]** |
| Transformer+RF | **96.27%[2]** | 39,160.2[5] | 33,602.0[5] | 32057088[5] |
| HPPI-Net | **96.70%[1]** | **143.9[1]** | **1312.4[2]** | 1173425[3] |

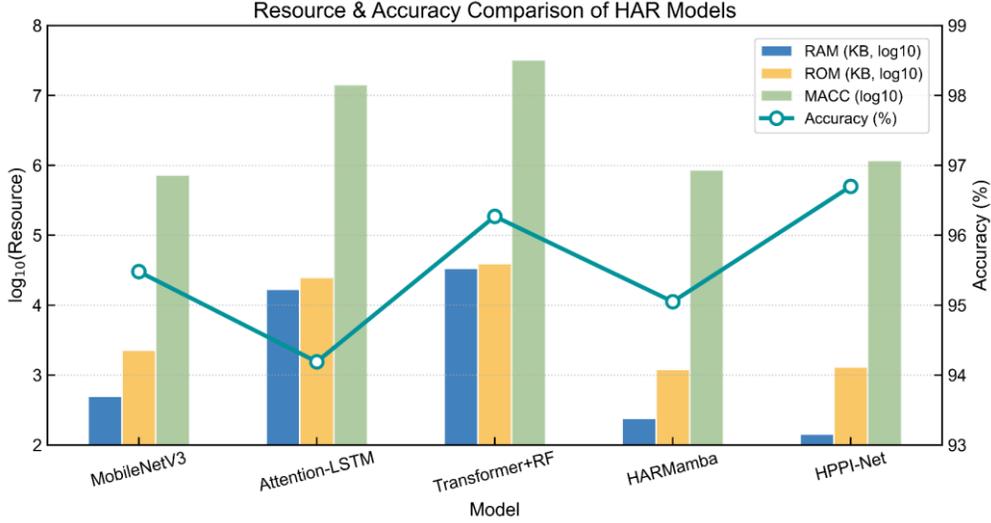

**Fig. 9.** Resource and Accuracy Comparison of each HAR model.

As summarized in **Table 4**, HPPI-Net achieves the highest classification accuracy (96.70%) while maintaining one of the lowest memory and compute footprints among all evaluated models. Compared with MobileNetV3, HPPI-Net improves accuracy by 1.22%, while reducing RAM usage by 71.2% and saving ROM by 42.1%. In contrast, Transformer+RF and Attention-LSTM both achieve reasonable accuracy (96.27% and 94.19%, respectively), but require two orders of magnitude more RAM (up to 39160 KiB) and ROM (up to 33602 KiB), rendering them unsuitable for embedded deployment.

Although HARMamba exhibits a lower memory footprint (RAM: 240.2 KiB, ROM: 1204.1 KiB), it suffers from a 1.65% drop in accuracy, which is critical in real-world HAR applications. In contrast, HPPI-Net preserves performance while still fitting within microcontroller-class memory budgets.

**Fig. 9** illustrates the trade-off between resource usage and accuracy. RAM, ROM, and MACC are shown on a $\log_{10}$ scale (bars), while accuracy is presented in percentage (line plot). HPPI-Net lies on the Pareto frontier. HPPI-Net lies on the Pareto frontier, indicating that no other model offers better accuracy at a lower resource cost. This establishes HPPI-Net as the most deployment-ready approach among all compared methods.

## 5. Conclusion

This work proposed HPPI-Net, a two–stage hierarchical-inference framework that integrates depthwise separable convolution with multi-spectral fusion (FFT,WT, and GT) and channel-level interpretability to reconcile recognition accuracy with the stringent resource envelope of microcontroller-class hardware. The lightweight FFT-based First_Layer classifier first performs an initial coarse discrimination and directs each window to either a dynamic-activity path or a stationary-activity path; only the dynamic branch invokes the more expressive PLMN module. Both branches share the identical DSC backbone, and their feature streams are subsequently concatenated via an attention-driven interpretability layer. This architecture preserves discriminative capacity and transparency while remaining within the tight computational and memory budgets characteristic of on-device inference.

Implemented on ARM Cortex-M4 MCU, HPPI-Net attains an overall accuracy of 96.70% while occupying only 22.3KiB of SRAM and 439.5KiB of flash. These savings arise from the systematic use of depthwise separable convolution, adapted from the MobileNet design paradigm to curtail multiply–accumulate operations, without adopting the full and resource-intensive MobileNet macro-architecture. Relative to a functionally equivalent MobileNetV3 baseline, HPPI-Net delivers an accuracy gain of 1.22% and simultaneously reduces SRAM by 71.2 % and flash by 42.1 %, while sustaining real-time HAR tasks on microcontroller. This blend of fidelity, interpretability, and ultralow memory footprint positions HPPI-Net as a compelling solution for forthcoming generations of smart-wearable platforms, condition-monitoring nodes, and other edge devices where both performance and on-chip resource austerity are paramount.